%% file: bare_adv.tex
\documentclass[10pt,journal,compsoc]{IEEEtran}

\usepackage{color, colortbl}
\usepackage{array}
\definecolor{Gray}{gray}{0.8}

\def\Heat{{\mathcal{H}}}
\def\Net{{\Phi}}
\newcommand{\Ne}[2]{\Net( #1 \, , \theta_{#2})}

\usepackage{graphics} 
\usepackage{epsfig} 
\usepackage{times} 
\usepackage{amsmath} 
\usepackage{amssymb}  
\usepackage{multirow}
\usepackage{textcomp}
\usepackage{comment}
\usepackage{lipsum}

\usepackage{balance}

\ifCLASSOPTIONcompsoc
  \usepackage[nocompress]{cite}
\else
  \usepackage{cite}
\fi
\ifCLASSINFOpdf
\else
\fi
\begin{document}
%
\title{A Transfer Learning approach to Heatmap Regression for Action Unit intensity estimation}

\author{Ioanna Ntinou, \and Enrique Sanchez, \and  Adrian Bulat, \and
        Michel Valstar, \and
        and~Georgios Tzimiropoulos
\IEEEcompsocitemizethanks{\IEEEcompsocthanksitem Ioanna Ntinou and Michel Valstar are with the Computer Vision Lab, School of Computer Science, University of Nottingham, NG8 1BB, UK. \protect\\
E-mail: \{ioanna.ntinou1,michel.valstar\}@nottingham.ac.uk
\IEEEcompsocthanksitem Georgios Tzimiropoulos is with the School of Electronic Engineering and Computer Science, Queen Mary University of London, E1 4NS , UK. \protect\\
E-mail: g.tzimiropoulos@qmul.ac.uk
\IEEEcompsocthanksitem Enrique Sanchez, Adrian Bulat and Georgios Tzimiropoulos are with Samsung AI Center
Cambridge, CB1 2RE, UK. 
\protect\\
E-mail: kike.sanc@gmail.com; adrian@adrianbulat.com \protect

}

\thanks{Manuscript submitted on April 14, 2020}}

%
%

\markboth{Submitted to IEEE Transactions on Affective Computing}%
{Ntinou \MakeLowercase{\textit{et al.}}: Title}
%



\IEEEtitleabstractindextext{%
\begin{abstract}
Action Units (AUs) are geometrically-based atomic facial muscle movements known to produce appearance changes at specific facial locations. Motivated by this observation we propose a novel AU modelling problem that consists of jointly estimating their localisation and intensity. To this end, we propose a simple yet efficient approach based on Heatmap Regression that merges both problems into a single task. A Heatmap models whether an AU occurs or not at a given spatial location. To accommodate the joint modelling of AUs intensity, we propose variable size heatmaps, with their amplitude and size varying according to the labelled intensity. Using Heatmap Regression, we can inherit from the progress recently witnessed in facial landmark localisation. Building upon the similarities between both problems, we devise a transfer learning approach where we exploit the knowledge of a network trained on large-scale facial landmark datasets. In particular, we explore different alternatives for transfer learning through a) fine-tuning, b) adaptation layers, c) attention maps, and d) reparametrisation. Our approach effectively inherits the rich facial features produced by a strong face alignment network, with minimal extra computational cost. We empirically validate that our system sets a new state-of-the-art on three popular datasets, namely BP4D, DISFA, and FERA2017.

\end{abstract}

}

\maketitle

\IEEEdisplaynontitleabstractindextext

%
\IEEEpeerreviewmaketitle


%
%
%
%

\input{content/introduction.tex}

\input{content/related_work.tex}
\input{content/heatmap_regression.tex}
\input{content/incremental.tex}

\input{content/implementation.tex}

\input{content/inhouse_evaluation.tex}
\input{content/comparison_soa.tex}

\section{Conclusion}

We have proposed a simple yet efficient approach for the problem of facial Action Unit intensity estimation: that of joint localisation and intensity estimation through heatmap regression. To accommodate the varying AU levels in the framework of heatmap regression, we modify the ground-truth heatmaps by changing their size and amplitude according to the corresponding AU intensity. Then, motivated by the similarities of our approach with these of the face alignment task, along with the fact that the task of face alignment is equipped with rich annotations, we reform the task of AU heatmap regression with an incremental learning approach. To do so, we incorporate to our setting a pre-trained facial landmark network that provides us with rich face related features across a variety of poses and illuminations. We conducted extensive experiments illustrating how the proposed approach systematically improves Intra Class Correlation (ICC) and thus achieve state of the art results on three benchmark datasets: FERA2015, DISFA and FERA2017. 

\appendices


\ifCLASSOPTIONcompsoc
  \section*{Acknowledgments}
\else
  \section*{Acknowledgment}
\fi

The work of Ioanna Ntinou was supported by the Horizon Centre for Doctoral Training, School of Computer Science, University of Nottingham. The work of Michel Valstar was co-funded by the NIHR Nottingham Biomedical Research Centre. 
\ifCLASSOPTIONcaptionsoff
  \newpage
\fi



%
\balance
\bibliographystyle{abbrv}
\bibliography{refs}


%





\end{document}

%% file: content/introduction.tex
\ifCLASSOPTIONcompsoc
\IEEEraisesectionheading{\section{Introduction}\label{sec:introduction}}
\else
\section{Introduction}
\label{sec:introduction}
\fi
\label{sec:intro}

\IEEEPARstart{A}{utomatic} facial expression analysis is important for detecting, recognising and interpreting the emotional state underlying a given facial image. One of the most common descriptors of facial expressions are the Action Units (AUs,~\cite{ekman02}). The Facial Action Coding System (FACS,~\cite{ekman02}) defines Action Units as atomic non-overlapping facial muscle actions that when combined in different configurations can describe any facial expression. There are 32 AUs in total. The Facial Action Coding System also establishes a six-point ordinal ranking of intensities which ranges from $0$ to $5$, with $0$ denoting the absence of a specific Action Unit, and $5$ referring to the maximum level of expressivity.

Action Units are in many ways inherently correlated: \textit{in time}, as facial actions vary smoothly within a sequence, in their \textit{co-occurrence}, as Action Units are often activated in certain meaningful combinations, and  \textit{spatially}, as they adhere to anatomically defined local and global geometric structure. In the field of automatic facial expression analysis, these correlations can indeed be regarded as a line of research, either alone or in combination with others. For example, co-occurrence correlation is exploited to perform joint prediction of multiple AUs, either through shared feature representations~\cite{zhang2014, zhu2014} or through methods that impose correlations among labels, usually by employing graphs~\cite{walecki16, walecki17}. Furthermore, a significant amount of works attempt to exploit
spatial correlations of AUs by extracting local representations in
the facial regions where AUs are known to produce appearance
changes, namely Regions of Interest (ROIs)  \cite{jaiswal2016, li2017b, li2017c}. Typically spatially-aware approaches employ a two-stage pipeline where facial landmarks, are firstly detected in order to define AU locations,  and then, local
features for each AU are extracted and, adaptation or fusion mechanisms are
introduced to jointly predict AU intensity levels. Our method significantly simplifies the aforementioned two-stage pipeline by localising AUs and estimating their intensity, while also modelling their inter-dependencies, in a single step.

\begin{figure}[tpb!]
    \centering
    \includegraphics[width=0.44\textwidth]{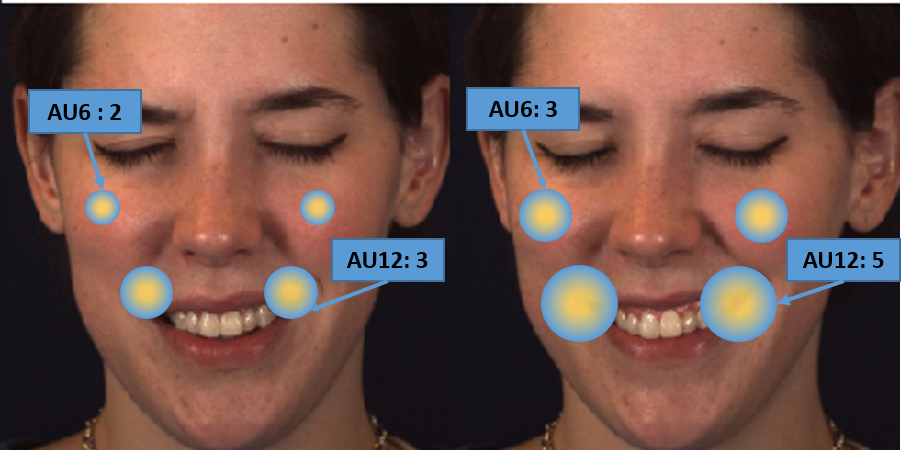}
    \caption{Facial AUs have strong spatial correlations and co-dependent structure. While the spatial correlation comes from their definition, we often find AUs that tend to co-occur. For instance, AU 6 (cheek raising), and AU 12 (lip corner pulling) are known to correlate. In this paper, we propose to jointly model their localisation and their intensity in a simple yet efficient Heatmap Regression manner, where the Heatmaps are chosen to depend on the corresponding AU intensities. }
    \label{fig:loc}
\end{figure}

\begin{figure*}[htp!]
    \centering
    \includegraphics[width=0.98\textwidth]{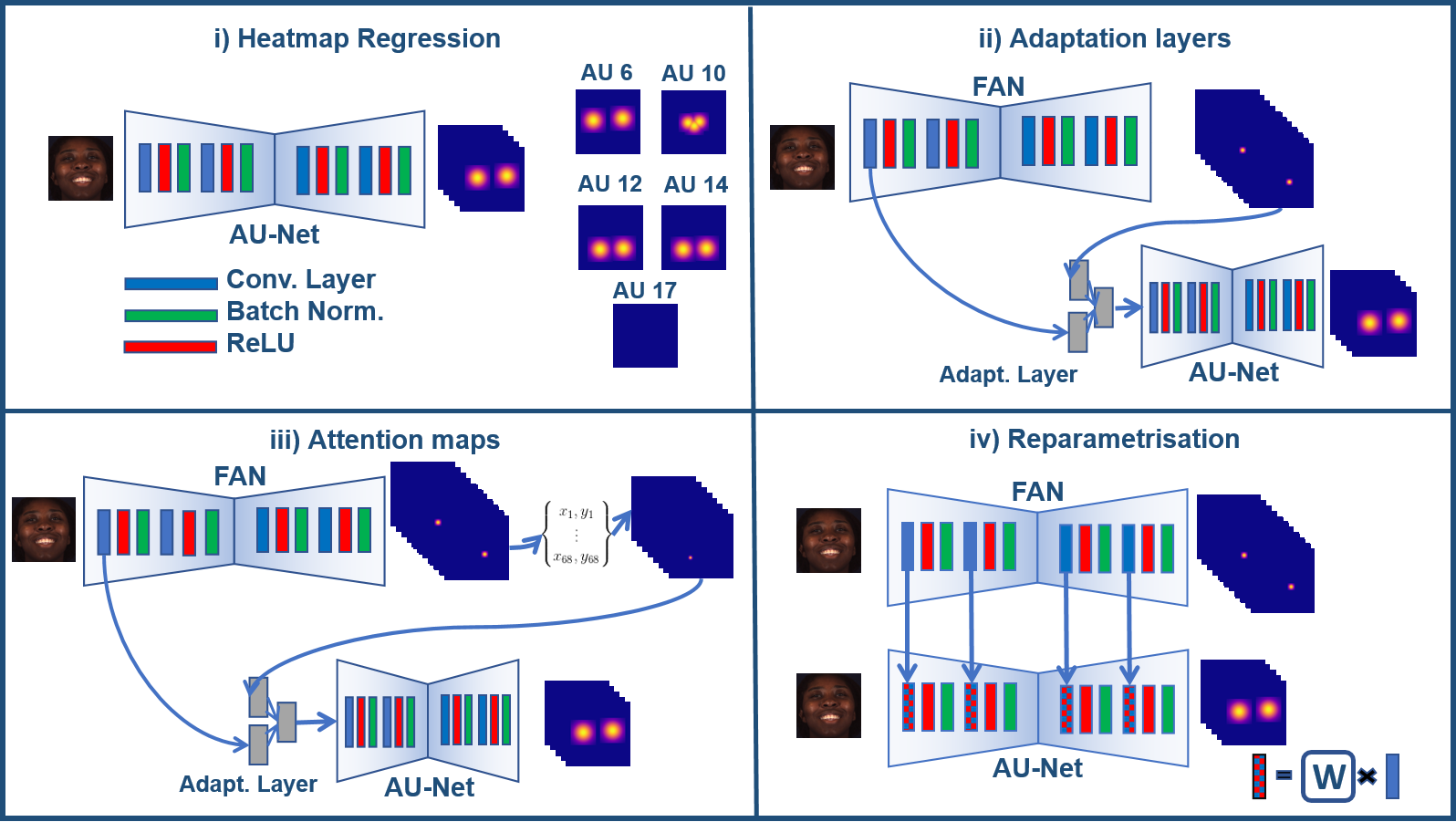}
    \caption{Overview of the main approaches compared in this paper. i) This paper poses the problem of AU intensity estimation in a Heatmap Regression framework, where the AU network, (referred for simplicity as \textbf{AU-Net}) is trained to jointly localise and estimate the intensity of Action Units. We then observe the similarities between this approach and that known to deliver state of the art results in the task of Face Alignment (\textbf{FAN}), and propose three alternatives to incrementally learn our AU-Net. In particular, we propose. ii) an approach where one can use the early features given by a FAN, along with the generated facial landmarks; iii) an approach where the facial landmarks are used to produce AU attention maps, to be fed to the AU-Net alongside with the early features of FAN and; iv) an approach where the original FAN is reparameterised using a small set of additional parameters that can be projected onto the weights of the filters of the FAN.}
    \label{fig:abstract}
\end{figure*}

In this paper, we firstly make the simple observation that AU recognition should be treated as localisation problem where the task is to both localise and classify the Action Units.
Motivated by this observation, we propose a new task that consists of jointly localising and estimating the intensity of Action Units, and we formulate this problem using Heatmap Regression. Heatmap Regression is arguably the most successful approach to landmark (keypoint) localisation~\cite{newell2016,bulat2017}. It boils down to a pixel-wise regression task that indicates the likelihood of a landmark being present at the corresponding spatial location. However, unlike landmarks, AUs can be present or absent in a given image with their intensity spanning from $0$ to $5$. Hence, their amplitude modelling cannot be treated with the standard probabilistic approach of Heatmap Regression (i.e. with the heatmaps relating to the confidence of a detected landmark). To overcome such limitation, we extend Heatmap Regression to include maps that are modelled according to the corresponding AU intensity. In particular, motivated by the fact that Action Units produce appearance changes around the facial region where they are known to occur, we propose to model the size of a heatmap (i.e. its amplitude and extend) according to the intensity of a given Action Unit. Under this setting, our idea boils down to a simple yet efficient Heatmap Regression approach. This approach is efficient in a way that not only merges the co-occurrence and the spatial correlation of AUs in a single task, but also in a way that bypasses all complexities associated to the typical two-stage pipeline, consisting of registration followed by local feature extraction and classification, often found in AU related works.

By jointly tackling the problem of AU intensity estimation and localisation using Heatmap Regression one could choose to dismiss the commonly required step of face alignment. Rather doing so, in this work we further propose to integrate this task into our system using transfer learning. In particular, on one hand it is known that AU annotations are scarce and hence it is difficult to train a system for AU recognition that can work well across all types of facial variability (e.g. facial pose, illumination, occlusion). On the other hand, there is a large pool of facial landmark annotations available for all types of facial variability. Since heatmap regression can be used to tackle both tasks, we investigate how and to what extent one can transfer knowledge from a network trained for landmark localisation into a network for AU localisation and intensity estimation. We explore several alternatives for transfer learning through a) fine-tuning, b) adaptation layers, c) attention maps, and d) reparameterisation (see Fig.~\ref{fig:abstract}). We show that our approach allows for robust AU modelling across a wide range of poses and illumination conditions.

This paper reformulates our previous manuscript on Action Unit intensity estimation using Heatmap Regression~\cite{sanchez2018} and devises a more robust approach using transfer learning. We show that this approach benefits from the robustness of the facial features given by a similar network trained for the task of facial landmark localisation, yielding state of the art results in three different datasets (FERA 2015, DISFA, FERA 2017). The contributions of this paper are as follows:
\begin{itemize}
    \item We propose to reformulate the problem of Action Unit intensity estimation in a way that absorbs their localisation. In particular, we are the first to propose to jointly localise and estimate the intensity of Action Units using Heatmap Regression. The use of variable size heatmaps allows the joint modelling of AUs localisation and intensity estimation in a single yet efficient way. 
    \item We propose the use of transfer learning to exploit the knowledge of a network trained for a similar task, that of face alignment, in a large-scale of images ranging a wide variety of poses, expressions, and illumination, conditions often hard to find in AU datasets. To this end, we explore several variants and identify an incremental learning approach which significantly reduces the number of weights to be trained, and increases robustness against different views. 
    \item We extensively validate our approach in three challenging datasets, namely BP4D~\cite{valstar2015}, DISFA~\cite{mavadati2013}, and FERA 2017~\cite{valstar2017}, yielding state of the art results with an approach that requires little complexity. 
\end{itemize}

%% file: content/related_work.tex
\section{Related Work}
\label{sec:related_work}

\noindent We firstly review the closely related work in the area of facial Action Unit detection and intensity estimation, and then we provide some insight into existing approaches to transfer learning. 

\subsection{Action Unit detection and intensity estimation}
\noindent Facial Action Unit modelling is a longstanding problem in Computer Vision and Affective Computing, that is often split into works that either target their \textit{detection}, i.e. estimating whether an AU is present or not on a given facial image~\cite{almaev13,almaev2015,Baltrusaitis2015,chu13,chu2017,ertugrul2019,he2017,jaiswal2016,jiang12,li2017b,li2017c, Niu_2019_CVPR,shao2018,wu16,yang2019}, or the more challenging task of estimating their intensity~\cite{eleftheriadis2016,jeni13,rudovic12,rudovic13a,rudovic13b,rudovic2015,sandbach13,tran2017,walecki16,walecki17}, given as a value that ranges from $0$ (i.e. absence of an AU), to $5$ (maximum intensity). Regardless the task, many methods partially share the underlying methodology: some works attempt to leverage co-occurrence and static dependencies among AUs~\cite{walecki17, ertugrul2019}, some exploit their geometric structure~\cite{zhao2016,  li2017c}, or their temporal correlation in time~\cite{jaiswal2016} and works that combine different means of correlation~~\cite{ming2015, li2017b,chu2019, jaiswal2016}. 

One of the most exploited means of AU correlation refers to their spatial structure. Action Units have a geometric structure, i.e. they are spatially correlated to specific facial regions. Early works on AU modelling were targeting the design of some handcrafted features that can inform about local appearance variations that are ultimately related to each AU~\cite{almaev13}. With the development of Convolutional Neural Nets this design was no longer needed, and other techniques to extract local features appeared. A simple approach to extracting local features is that of \cite{jaiswal2016}, where the face is first registered according to some landmark detection, and each part is cropped independently. Then, CNN-based features can be extracted independently. In a similar fashion, \cite{zhao2016} proposed to incorporate an intermediate region-specific layer to a CNN to extract separate features at different facial sub-areas, while \cite{li2017c} incorporated to a pre-trained CNN two extra layers - coined as the enhancing and the cropping layer - to enforce the network pay more attention to spatial regions with high AU correlation. With such locally-based modelling, \cite{chu2017} proposed to introduce a temporal model in a hybrid manner, to jointly exploit the local and temporal correlation of AUs. Building on top of region layers, \cite{li2017b} introduced the CNN-based Region of Interest (ROI) detection, which was then incorporated into the local modelling of AUs. In particular, \cite{li2017b} proposed independent ROI networks to learn separate filters for different facial regions, which were later used to feed a fully-connected LSTM network to also exploit the temporal correlation. More recently, \cite{tran2017} proposed to model AU regions by incorporating a Variational Autoencoder framework (VAE,~\cite{kingma2013auto}), and \cite{yang2019} combined a 2D with 3D CNN for frame-level AU detection at an attempt to leverage spatio-temporal dependencies. All of these works require, however, a good pre-processing step that consists of registering the input image according to some detected facial landmarks. In this paper, we observe that both tasks can be performed together in a rather simple way. While the methodology proposed in this paper is completely new, some  works have attempted to jointly detect facial landmarks and perform AU modelling in a unified framework. In an early work, \cite{wu16} proposed an iterative framework whereby a cascaded regression approach was used to detect facial landmarks, and where a Restricted Boltzmann Machine was used to detect the Action Units. Recently, \cite{shao2018} proposed to jointly perform facial landmark localisation and AU intensity estimation through a hierarchical, multi-scale region learning pipeline that employs attention maps refinement to ease the learning process. In \cite{Niu_2019_CVPR} the landmarks are instead used to regularise the features extracted by a CNN, towards driving these to be person-specific. Both \cite{Niu_2019_CVPR} and \cite{shao2018} observe that landmarks carry over important information, either to regularise the features or to generate attention maps. In a more recent approach, \cite{shao2019} proposed the use of attention maps that are landmark-free, showing how locally-based features can better model the AU occurrence. In this paper, we propose a rather less computationally expensive method that can deal with estimating the AU intensity by first re-formulating the problem in a way that includes their localisation, and then by incorporating the rich features acquired by a network trained to detect facial landmarks. Our approach offers a significantly less complex yet efficient method that delivers state-of-the-art results in the more challenging problem of estimating the intensity of Action Units.

Finally, it is worth mentioning the recent appearance of methods that work on a weakly supervised or even unsupervised manner~\cite{li2019,Zhang_2018_CVPR,Zhang_2018_CVPR_BN,wang_2019,Zhang_2018_CVPR_weakly,Zhang_2019_ICCV,niu2019multi}. While these works have shown some interesting advances, they are still behind the performance achieved by fully supervised methods. Although it is out of the scope of this paper, the low computational complexity of our method suggests that it could also be a good approach for learning with scarce labels. We leave this problem for future work.

\subsection{Transfer Learning}

The goal of transfer learning, often found in the literature as incremental learning, is to adapt the knowledge acquired for a strong task for which a large pool of labels is available to learn a set of potentially unrelated tasks~\cite{rosenfeld2018incremental, rebuffi2017learning, rebuffi2018efficient}. One of the simplest approaches to transfer learning consists of fine-tuning~\cite{huh2016makes}, often used in face analysis works. For example, it is a common practice to initialise a network with the weights of the pre-trained VGG-Face2~\cite{cao2018vggface2}, trained with thousands of images for the task of face recognition. Some examples of this can be found in~\cite{li2017c,li2017b,Niu_2019_CVPR}. 
Other works have proposed to add knowledge incrementally, i.e by extracting features from a model trained on a specific task and use them to train another model for a new task. This method advances over fine-tuning in the sense that a model is trained for a new task without forgetting old representations. An example is the progressive networks~\cite{rusu2016progressive}, or the adaptive filters~\cite{rebuffi2017learning,rebuffi2018efficient}.

In this paper, we explore different alternatives for transfer feature learning, as well as an approach based on network reparameterisation, which consists of applying a transformation over the existing weights of the pre-trained network. Network reparametrisation is often found in incremental learning approaches, where new tasks are added sequentially to a strong core. For instance, \cite{kossaifi2019t} applies a tensor decomposition that allows each of the tensor dimensions to adapt a new task. A simple approach that has been shown to perform  well in practice consists of projecting the given weights in a convolutional network with a simple projection matrix, learned over the new task~\cite{rosenfeld2018incremental}. This approach was also recently applied to the unsupervised adaptation of object landmark detectors~\cite{sanchez2019}. We observe that we can use such a framework in a supervised manner to transfer the knowledge of a face alignment network to our proposed AU intensity estimation network. We observe that this approach offers significant computational benefits whilst yielding state of the art results.

%% file: content/heatmap_regression.tex
\section{Joint AU localisation and intensity estimation}
\label{sec:original}

\begin{figure}[t!]
    \centering
    \includegraphics[width=0.40\textwidth]{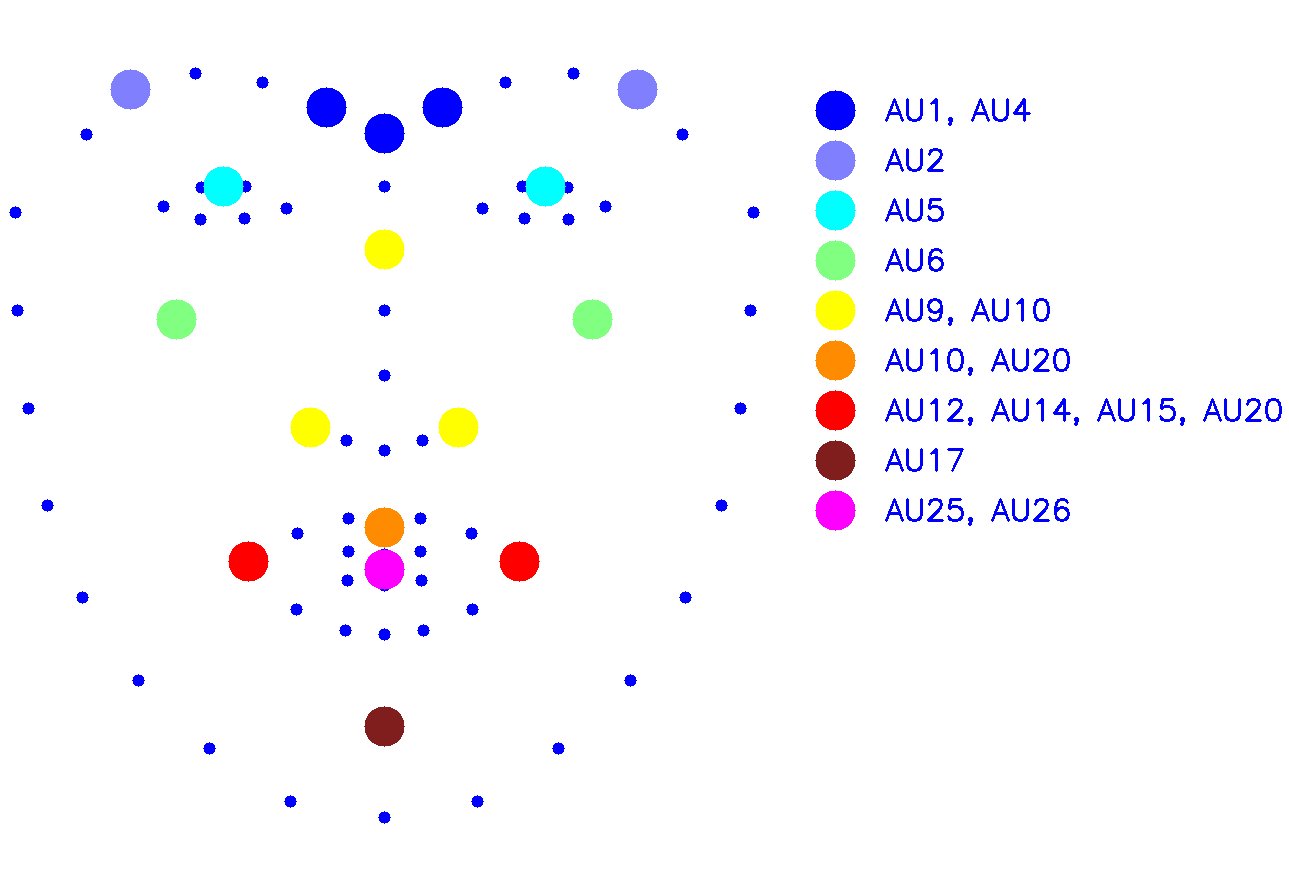}
    \caption{
    Dots correspond to the facial landmarks and circles correspond to the location of various AUs. First, facial landmarks are used to define the location of the AUs. Then, Gaussians with varying amplitude are generated in the spatial location of AUs. Some AUs share the same spatial location and hence their location is defined by the same landmarks. For example, AUs  9 and 10 share two Gaussians, but each of them will activate according to their intensity. See Table~\ref{table:correspondences} for all the correspondences between AU spatial locations and facial landmarks.}
    \label{fig:truth}
\end{figure}

\noindent In this Section, we present a novel approach to AU intensity estimation using heatmap regression. The main novelty of this approach lies in the fact that heatmap regression allows joint localisation and intensity estimation of the AUs, as the machine learning task gains a spatial aspect. By using an encoder-decoder approach, we are able to gather features at different spatial levels to yield a dense pixel-wise prediction, facilitating inference, and allowing the network to learn both the spatial relation and co-occurrence of AUs. Our approach differentiates from previous works that apply multi-task learning through joint estimation of AU intensities and landmark localisation, in the sense that AUs' spatial relation and co-occurrence is inherently embedded in the heatmap regression method. In addition, Heatmap Regression does not make use of attention maps to predict the score, it \textit{just} regresses the heatmaps. Our approach offers significant advances: it yields state of the art results without requiring any complex pre-processing or face alignment, thus effectively reducing the computational cost of inference. 

\subsection{Problem formulation}

\noindent Our goal is to train a network that can predict the intensity of a set of AUs. To do so, we propose to reformulate the training process in a way that makes the network jointly detect the location of the AUs, as well as their intensity. This way, we can formulate the joint problem using Heatmap Regression, thus relaxing the training process. This relaxation comes from the fact that Heatmap Regression boils down to local pixel-wise regression, making the network penalise local errors more efficiently, rather than in a global way, as is the case of direct regression. 
While some works use the facial landmarks to perform Multi-Task learning or generate some attention, we want to formulate a joint training, where the set of images and AUs is augmented by the locations of the latter. Formally, let $\mathrm{D} = \{{\bf I}_i \in \mathbb{R}^{3\times W \times H}\}_{i=1}^N$ be a set of $N$ RGB $W\times H$ images, for which the corresponding AU intensities ${\bf a}_i = (a_{i,1}, \dots , a_{i,{N_{aus}}})$ are known, with $N_{aus}$ the number of annotated AUs in the dataset. Our first goal is to extend the training annotations with a set of locations ${\bf p}_i = ({\bf p}_{i,1}, \dots, {\bf p}_{i,N_{aus}})$, where each ${\bf p}_{i,j} \in \mathbb{R}^{2 \times N_{i,j}}$ corresponds to the $x,y$ coordinates of the $N_{i,j}$ locations where an AU $j$ is known to produce appearance changes for that particular AU when it is activated. Each AU will often be placed at two locations, symmetric with respect to the facial axis of symmetry. Some of the AUs will also produce changes at a third location, placed along the symmetry axis. Given that there are no annotations available to place the ground-truth locations, we place them using a priori knowledge about their location and with respect to the $68$ facial landmarks used for landmark localisation. The positions of the AUs as derived from the given landmarks is given in Fig.~\ref{fig:truth}. The exact correspondences are illustrated in Table~\ref{table:correspondences}. After having generated the AU positions, our training set is now defined as $\mathrm{D} = \{ {\bf I}_i, {\bf a}_i, {\bf p}_i \}_{i=1}^N$.

\begin{table}[htb!]
\caption{Correspondences between landmarks and AUs. In our setting each AU can be described with the activation  of up to three Gaussians on the right, left and centre of the face. Landmark indexing is as noted by~\cite{sagonas2016}}
\label{table:correspondences}
\begin{center}
\begingroup
  \begin{tabular}{|p{0.001\textwidth}>{\raggedleft} l|*{3}{c} |c|}
  \hline
  \rowcolor{Gray} & AUs & \multicolumn{3}{|c|}{Facial Landmarks}\\
  \rowcolor{Gray}
		&  & \multicolumn{1}{|c}{Left} & \multicolumn{1}{c}{Right} & \multicolumn{1}{c|}{Centre} \\
	\hline \hline
 & AU 1 & 21 & 22& 21,22,27  \\ 
 \hline
 & AU 2 & 18 & 25 & - \\
 \hline
 & AU 4 & 21 & 22& 21,22, 27  \\
 \hline
 & AU 5 & 37,38 & 43,44 & -\\
 \hline
 & AU 6 & 1,41,31& 15,46,35 & - \\ 
 \hline
 & AU 9 &31 &35 &28  \\
  \hline
 & AU 10 & 31 & 35 & 51  \\
  \hline
 & AU 12 & 48 & 54 & -  \\
  \hline
 & AU 14 & 48 & 54 & -  \\
  \hline
 & AU 15 & 48 & 54 &- \\
  \hline
 & AU 17 & 57 & 8 & - \\
  \hline
 & AU 20 & 48& 54 & 51\\
  \hline
 & AU 25 & - &  -&61,64  \\
 \hline
 & AU 26& - &  - &61,64\\
 \hline
\end{tabular}
\endgroup
\end{center}
\end{table}

\subsection{Heatmap Regression}
\label{ssec:heatmaps}
Once the training set has been augmented with the AU locations, we can define the training procedure. Inspired by the success of Heatmap Regression for facial landmark localisation, we propose to formulate the training problem in a similar way. However, locating the AUs only, would not solve the problem of estimating their intensity, and thus we need to accommodate the latter into the localisation problem. To do so, we propose to attach each AU to a corresponding heatmap. Each heatmap will contain one, two or three Gaussians, according to the number of points defined in ${\bf p}_i$ (see Table~\ref{table:correspondences} for all different correspondences). Following existing works in Heatmap Regression, we will work with heatmaps that are quarter the size of the input image, i.e. ${\bf p}_{i,j} \leftarrow 0.25 {\mkern -0.5mu \cdot \mkern -0.5mu} {\bf p}_{i,j}$. Each Gaussian $g_{i,j}$ is defined as a 2D map having the same size as that of the input image, where the value at each position ${\bf x}$ is given by:
\begin{equation}
   g_{i,j}({\bf x}) =  a_{i,j} \exp{\Big( \frac{\|{\bf x} - {\bf p}_{i,j}\|}{a_{i,j}^2 }\Big) }
\end{equation}
 with $a_{i,j}$ the intensity of the $j$-th AU for image $i$. The heatmap for AU $j$ is thus defined as $\Heat_{i,j}({\bf x}) = \underset{k}{\max} \, g_{i,j}({\bf x})$. Under this representation, the heatmaps form an $N_{aus} \times W/4 \times H/4$ tensor, with $N_{aus}$ the number of AUs, and $W$ and $H$ the width and height of the images, respectively ($256$ in our setting). 

Our goal is then to train a network $\Ne{\cdot}{}$ that, given an image ${\bf I}$, regresses a set of heatmaps $\Heat \in \mathbb{R}^{N_{aus} \times W/4 \times H/4}$. The network is parametrised by the parameters $\theta$. The learning is formulated in a standard heatmap regression fashion, i.e. as finding the weights $\theta$ that minimise the squared loss between the output and the ground-truth maps:
\begin{equation}
    \theta^* = \underset{\theta}{\arg \min} \sum_{i=1}^N \| \Heat_i - \Ne{{\bf I}_i}{}  \|^2
\end{equation}
\subsection{Network}
\label{ssec:network}
\begin{figure}[h!]
    \centering
    \includegraphics[width=0.30\textwidth]{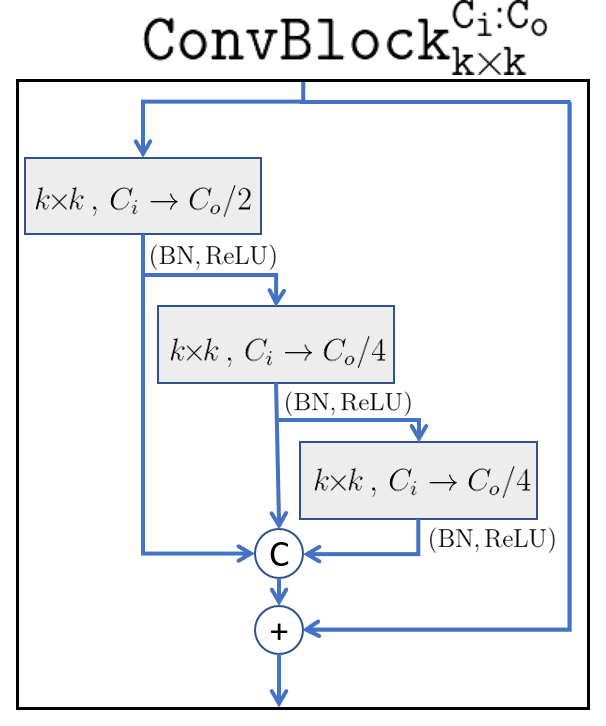}

    \caption{Convolutional Block, main building block in all the networks in this paper. The block receives an input tensor with $C_i$ channels, and produces an output tensor with the same spatial resolution and $C_o$ output channels. The kernel filter is $k\times k$}
    \label{fig:convblock}
\end{figure}

For the sake of clarity, we introduce the network description and training herein, as it will constitute what we refer to as the~\textit{backbone} throughout the next section. In particular, the network $\Net$ follows a similar architecture as that of the Face Alignment Network (FAN,~\cite{bulat2017}), with small differences. In our setting, the network $\Net$ receives an input image ${\bf I} \in \mathbb{R}^{3 \times 256 \times 256}$, and first applies a downsampling convolutional $7\times7$ filter to it, halving its resolution and increasing the number of channels to $64$. Then, a set of $3$ Convolutional Blocks (referred to as $\mathtt{ConvBlock}$~\cite{bulat1027b}, see Fig.~\ref{fig:convblock}), are used to bring the number of channels to $128$ and the spatial resolution to $64\times64$. We will refer to these layers as $\mathtt{conv2}$, $\mathtt{conv3}$, and $\mathtt{conv4}$, respectively. Then, the features after the $\mathtt{conv4}$ layer are fed into a single Hourglass network~\cite{newell2016} (Fig.~\ref{fig:convblock}), which is an encoder-decoder network, composed of several $\mathtt{ConvBlock}$ and skip connections that aggregate the features at different scales. While \cite{newell2016,bulat2016} used a set of $4$ stacked Hourglass with $256$-channel $\mathtt{ConvBlock}$, we opt for a lighter version that consists of a single Hourglass with $\mathtt{ConvBlock}$ of $128$ channels. The output of the Hourglass is finally passed through an extra $\mathtt{ConvBlock}$ and a convolutional layer to bring the number of channels to the target $N_{aus}$ described above, i.e. it outputs the desired $N_{aus} \times 64 \times 64$ tensor. With such a lightweight model, the network comprises only $\sim1.6M$ parameters. We empirically validate that such a simple network yields competitive results whilst being computationally efficient. 


\subsection{Inference}
Inference in this setting is straightforward. To get the AU intensities, one simply needs to crop the face image according to some face detection and forward it to the trained network. The network returns a set of heatmaps, from which the AU intensities can be inferred by just finding the maximum of each map. Note that our method does not require to register the face image before inserting it to the network. 

\begin{figure}[h!]
    \centering
    \includegraphics[width=0.48\textwidth]{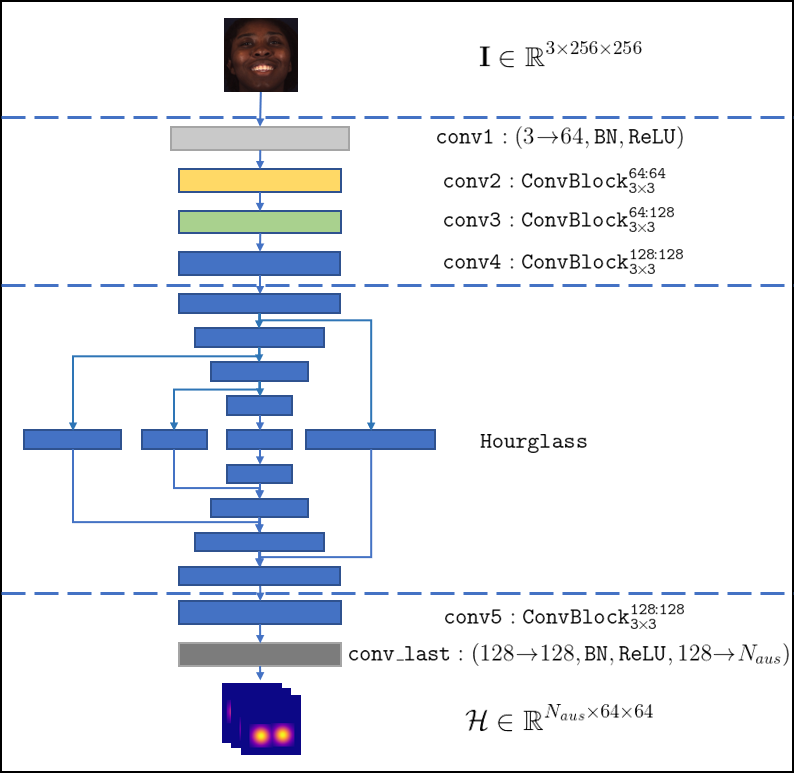}
    \caption{Main pipeline used for Heatmap Regression. The network receives an input image ${\bf I}$, and after passing it through several convolutional layers, it is fed into an Hourglass network. All the modules in the Hourglass are Convolutional Blocks as depicted in Fig.~\ref{fig:convblock}. The variable block size corresponds to halving, or doubling, the spatial resolution. The downsampling is done through max pooling after each Convolutional Block, whereas the upsampling is done through bilinear interpolation. }
    \label{fig:mainnet}
\end{figure} 

%% file: content/incremental.tex
\section{Incremental Heatmap Regression for AU localisation}
\label{sec:incremental}

\begin{figure}[tp!]
    \centering
     \includegraphics[width=0.47\textwidth]{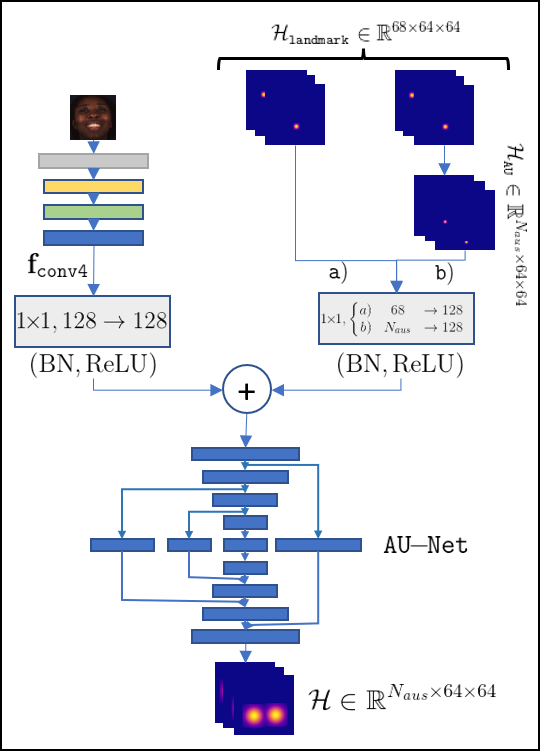}
    \caption{Proposed feature adaptation from FAN to the AU estimation model. We study two alternatives, where the early features coming from the $\mathtt{conv4}$ layer of the FAN are combined with a) the landmark heatmaps produced by the own FAN (Sec.~\ref{ss:feature_injection}), and b) a set of AU attention maps computed from the landmark heatmaps (Sec.~\ref{ss:attention_maps}). The features and the heatmaps are in both cases forwarded to a $1\times1$ filter to bring the number of channels to $128$ and ease the fusion. The corresponding outputs are added and sent to an Hourglass-like network (AU-Net), similar to that defined in Sec.~\ref{ssec:network}, with kernel filters set to $1\times1$.}
    \label{fig:connection}
\end{figure}

Using heatmap regression for AU recognition and localisation allows us to make use of the great progress that we have witnessed recently for the problem of facial landmark localisation. More specifically, we propose to transfer knowledge from a network trained for face alignment with hundreds of thousands in-the-wild images spanning a large set of poses, expressions, and illumination into the proposed network for AU intensity estimation and localisation. This in turn allows us to overcome to some extent the limitations of existing AU datasets related to facial variability (e.g. number of subjects, facial pose, occlusions etc.).

In contrast to the previous works that have attempted to exploit the correlation between facial expressions and localisation~\cite{wu16,shao2018} through a multi-task learning framework, we propose to use transfer learning to learn AU intensities from rich facial features retrieved from a pre-trained face alignment model.

The first and simplest approach to the proposed transfer learning approach consists of fine-tuning the pre-trained network for the target task. Besides fine-tuning, we propose and explore three different alternatives to accomplish the task of transfer learning, which are described below. First, we briefly describe the architecture and pre-training of the face alignment network (Sec.~\ref{ssec:pretraining}) and then, we explain how we fine-tune an AU estimation model from a face alignment one (Sec.~\ref{ss:finetuning}). Finally, we describe our three alternatives, namely that of~\textit{adaptation layers} (Sec.~\ref{ss:feature_injection}),~\textit{attention maps} (Sec.~\ref{ss:attention_maps}) and~\textit{network reparametrisation} (Sec.~\ref{ss:projection_matrix}). In what follows, we will refer to the face alignment network as \textbf{FAN}, whereas the corresponding part of the network targeted with the AU heatmap regression will be referred, for simplicity, as \textbf{AU-Net}. 

\subsection{FAN pre-training}
\label{ssec:pretraining}
 Our in-house implementation of FAN follows that of the AU-Net described in Sec.~\ref{ssec:network}. The output of the network is a set of $68$ heatmaps, each corresponding to a single landmark. The FAN is trained for $80$ epochs on LS3D-W \cite{bulat2017} training set which is the largest and most challenging facial landmark dataset to date (approximately $230,000$ images). The network yields a validation accuracy of $9.03$ point-to-point Euclidean distance, on par with that reported in ~\cite{bulat2017}.

\subsection{Method 1: Fine - tuning}
\label{ss:finetuning}
Our first and simplest approach to transfer learning consists of fine-tuning the pre-trained network. In particular, we observe that one can depart from the face alignment network and fine-tune it for the proposed AU localisation and intensity estimation by using a small learning rate. This will allow the network to slightly move from a very similar problem (that of facial landmark localisation) to our target task. We experimentally validate that, in line with existing works that suggest fine-tuning as a strong adaptation mechanism, such a simple technique already improves performance over training the network from scratch.

\subsection{Method 2: Adaptation layers}
\label{ss:feature_injection}
Our second approach to incremental learning consists of~\textit{transferring} the features generated from the Face Alignment Network (FAN) to a second network, targeted with producing the Action Unit heatmaps. In this paper, we conjecture that the early features produced by a strong FAN provide with rich facial representations, and we thus propose to~\textit{inject} this knowledge into a second learnable network. In addition to the early features, we also inject the produced heatmaps, as these are nothing but a geometric representation of the face. The heatmaps consolidate the spatial configuration of~\textit{all} landmarks, and hence encode information regarding location, pose, shape and expression of a face in an image. Finally, given that they are probabilistic maps, they provide both coordinate and confidence information which can be useful for understanding spatial context and modelling part relationships. Overall, we posit that the generated landmark heatmaps encode rich facial geometry representations that could operate as an attention mechanism that drives focus on regions of the face that are very informative for the task of AU prediction. Hence, it is reasonable to attempt to incorporate to the new task of AU estimation, this rich facial geometry information from face alignment. We study the impact of these heatmaps on the task of AU intensity estimation through several ablation studies in Sec.~\ref{sec:evaluation}.

The AU network has a similar structure than that of the Hourglass described in Sec.~\ref{ssec:network}. However, rather than using as input the facial image used to extract the facial landmarks, we use as input a combination of the features produced after the $\mathtt{conv4}$ block of the FAN, and the produced heatmaps ($68$). 

In order to inject the early features and produced heatmaps into the AU network, we use an adaptation layer, as depicted in Fig.~\ref{fig:connection}, a). This adaptation layer is composed of a branch that processes the early features coming from the FAN, and another branch that processes the generated landmark heatmaps. Let $\Heat \in \mathbb{R}^{68 \times 64 \times 64}$ be the output heatmaps corresponding to the facial landmarks, and let ${\bf f}_{\mathtt{conv4}} \in \mathbb{R}^{128 \times 64 \times 64}$ be the features from the $\mathtt{conv4}$ layer of the FAN. In order to integrate these two tensors, we apply to each of them a $1\times1$ filter, followed by a Batch Normalisation layer and a ReLU activation layer. The $1\times 1$ filters produce, for both cases, a $128 \times 64 \times 64$ tensor. The output of both branches are then added and sent to the AU network. The AU network then receives the combined features, and passes it through an Hourglass network with all filters set to have a kernel size of $1$, rather than the $3$ of the original FAN network. The output of this network is then a set of $N_{aus}$ heatmaps. The training is done through the classical heatmap regression depicted in Sec.~\ref{sec:original}. With the $1\times1$ filters, and the removal of the first convolutional blocks, the new network comprises only $\sim1M$ parameters.

\subsection{Method 3: Attention maps}
\label{ss:attention_maps}
A different alternative consists of generating~\textit{attention maps} from the generated heatmaps. This approach is depicted in Fig.~\ref{fig:connection}, b). In particular, it is important to recall, from Sec.~\ref{sec:original}, that at training time the target heatmaps, in the original setting, are located according to the ground-truth landmarks. In other words, there is a clear relation between the facial landmarks and the location of the Action Unit heatmaps. Thus, if we are to transfer the knowledge from the FAN network to the AU network, it is natural to explore the use of attention maps, generated from the heatmaps produced for the facial landmarks. 

In this setting, we use the corresponding heatmaps from the FAN, and extract the corresponding landmarks by applying an~\textit{argmax} operator. Then, using the method described in Sec.~\ref{ssec:heatmaps}, we generate a new set of heatmaps $\hat{\Heat} \in \mathbb{R}^{N_{aus} \times 64 \times 64}$. The heatmaps $\hat{\Heat}$ are then forwarded to the corresponding branch described above. However, it is worth noting that we do not have the ground-truth labels to produce a set of heatmaps that vary according to the Action Unit intensities. Instead, we are interested in generating an attention map, i.e. a heatmap that \textit{locates} the Action Units, without regarding to whether a given AU is actually present or not. To this end, the heatmaps in $\hat{\Heat}$ are generated using a fixed intensity $a = 1$.

Note that the attention maps are generated on the fly i.e  during training the generated heatmaps from the FAN are used to generate the attention maps. After having acquired the attention maps we follow the process described in Sec.~\ref{ss:feature_injection}. The new network is again $\sim1M$ parameters. 

\subsection{Method 4: Reparametrisation of FAN}
\label{ss:projection_matrix}

A different alternative consists of using the reparametrisation approach of \cite{rosenfeld2017, sanchez2019}. A visual representation is depicted in Fig.~\ref{fig:abstract}, iv).
In particular, we depart from the FAN network $\Ne{\cdot}{\text{FAN}}$, with parameters $\theta_{\text{FAN}}$, trained to detect facial landmarks, as described in Sec.~\ref{ssec:pretraining}. We now wish to adapt the model $\Net$ for the task of AU intensity estimation using heatmap regression, to yield a new set of weights $\theta_{\text{AU-Net}}$, so that $\Ne{\cdot}{\theta_{\text{AU-Net}}}$ would produce the desired AU heatmaps. As previously mentioned, $\Net$ is chosen to be an Hourglass, which is uniquely parametrised by convolutional and batch normalisation layers. The network $\Net$ is modified to return $N_{aus}$ heatmaps, rather than the $68$ heatmaps of facial landmarks, i.e. $\Ne{\cdot}{{\theta_{\text{AU-Net}}}}$ replicates the same structure for all layers but the very last one. Under this setting, the adaptation method proposed in \cite{sanchez2019} boils down to reparameterising the convolutional layers by learning a series of weights that are projected onto the original filters, to yield a new set of weights for the target task. Let us denote the weights of the convolutional layer $L$ of the original network as $\theta^L_{\text{FAN}} \in \mathbb{R}^{C_{in} \times C_{out} \times k \times k}$, with $C_{in}$ and $C_{out}$ the number of input and output channels, respectively, and $k$ the kernel size. Then, following \cite{sanchez2019} we use the following reparametrisation of the weights $\theta_L$:
\begin{equation}
    \theta^L_{\text{AU-Net}} = \mathbf{W}^L \times_{1} \theta^L_{\text{FAN}}
\end{equation}
where $\mathbf{W}^L \in \mathbb{R}^{C_{out} \times C_{out}}$ is the learnable projection matrix, and $\times_{n}$ denotes the $n$-mode product of tensors. The set of weights $\theta^L_{\text{AU-Net}}$ are of the same size than those of $\theta^{L}_{\text{FAN}}$, and can thus be replaced into the original network. Then, the learning is formulated in a heatmap regression fashion, although now the weights $\theta^{L}_{\text{FAN}}$ remain frozen, and only the weights $\mathbf{W}^L$ are to be learned. This approach, besides advancing the field of unsupervised adaptation, offers significant computational savings, as now the learnable weights have only $C_{out}^2$ parameters, contrary to the $C_{out} \times C_{in} \times k^2$ original set of parameters. Considering that the majority of filters in the Hourglass are of $k = 3$, the computational saving is, for $C_{in} = C_{out}$, about $9$ times the number of parameters. We observe that, while the original FAN network $\Net$ comprises $\sim 1.6M$ parameters, the new set of learnable weights $\mathbf{W}^L$ reduce to only $\sim 130K$ parameters. This method allows to efficiently transfer the knowledge from the pre-trained network to the target one.

%% file: content/implementation.tex
\section{Training and implementation details} 
\label{ssec:implementation}
In the following Sections, we evaluate each of the discussed alternatives for transfer learning, and compare them to training a model from scratch, as presented in Sec.~\ref{sec:original}. We then compare our method against existing works reporting AU intensity estimation. Note that, while we show qualitatively how our method is capable of localising the Action Units across a wide span of poses and expressions, we are primarily interested in demonstrating the superiority of our method at the task of estimating the intensity of AUs, and we are not interested in the precise localisation error.

\begin{table*}[htbp!]
\caption{Evaluation of different incremental learning methods on FERA 2015 development partition and DISFA. Bold numbers indicate best performance. \vspace{-15pt}
}
\label{table:evaluation_fera2015_disfa}
\begin{center}
\begingroup
\setlength{\tabcolsep}{5.5pt}
\resizebox{\textwidth}{!}{%
\begin{tabular}{|p{0.001\textwidth}>{\raggedleft} l|*{5}{c}|c|*{12}{c}|c|}
  \hline
  \rowcolor{Gray} & Dataset & \multicolumn{6}{c|}{FERA2015} & \multicolumn{13}{c|}{DISFA}\\
  \rowcolor{Gray}
		& AU	& \multicolumn{1}{c}{6} & \multicolumn{1}{c}{10} & \multicolumn{1}{c}{12} & \multicolumn{1}{c}{14} & \multicolumn{1}{c|}{17} & Avg. & \multicolumn{1}{c}{1} & \multicolumn{1}{c}{2} & \multicolumn{1}{c}{4} & \multicolumn{1}{c}{5} & \multicolumn{1}{c}{6} & \multicolumn{1}{c}{9} & \multicolumn{1}{c}{12} & \multicolumn{1}{c}{15} & \multicolumn{1}{c}{17} & \multicolumn{1}{c}{20} & \multicolumn{1}{c}{25} & \multicolumn{1}{c|}{26} & Avg.\\
	\hline \hline
	\multirow{6}{*}{\rotatebox{90}{\noindent ICC}} 
    & Scratch &  0.75 & {\bf 0.77} &0.85  &0.48  & 0.59 & 0.69 & \textbf{0.59 }& 0.56 & 0.74 & 0.36 & 0.53& 0.52 & 0.78 & 0.40 & 0.38 & 0.13 & 0.91 & 0.53  & 0.53 \\ 
    & Fine-tuning & 0.78 & 0.75 &0.88  &0.46  &0.66 &  0.71& 0.57 & \textbf{0.57} & 0.74 & 0.40 & 0.52& 0.50 & 0.80 & 0.51 & 0.42 & 0.16 & 0.93 & \textbf0.64 & 0.56 \\ 
    & Reparametrisation &  0.78  & 0.75 &0.86 & 0.49 & 0.63 & 0.70 &  0.54 & 0.51 & 0.68 & 0.40 & 0.49 & 0.49& 0.80 & 0.40& 0.36 & \textbf{0.23} & 0.91&  \textbf{0.69}& 0.54\\
    & Random Backbone & 0.75 &  0.72  & 0.85 & 0.41 & 0.47 & 0.64 & 0.31& 0.21 & 0.60 & 0.29 & 0.50& 0.46 & 0.77 & 0.36 & 0.40 & 0.14 & 0.88 & 0.61 &0.46 \\  
    & Attention maps& 0.78 & 0.73 & \textbf{0.89} & 0.49 & 0.67 & 0.71 & 0.51 & 0.50 &\textbf{ 0.77} & \textbf{0.47}  & \textbf{0.60} &0.50  & 0.82  &0.43 &\textbf{0.45}  &0.21  &0.93 & 0.62 & 0.57 \\ 
     & ResNet-18 & 0.74 & 0.75 & 0.85 & 0.40 & 0.52 & 0.65 & 0.41 & 0.35& 0.63& 032& 0.49& 0.48 & \textbf{0.83} & 0.31 & 0.32 & 0.20 & 0.84 & 0.50& 0.47 \\  
    & Adaptation layers & \textbf{0.79} & 0.76 & 0.84 & \textbf{0.52} & \textbf{0.67} & \textbf{0.72} & 0.56 & 0.52 & 0.75 & 0.42 & 0.51 & \textbf{0.55} & 0.82 & \textbf{0.55} & 0.37 & 0.21 & \textbf{0.93} & 0.62 & \textbf{0.57}  \\ 
    \hline \hline
	\multirow{6}{0.1cm}{\rotatebox{90}{MSE}} 
    & Scratch  & 1.00 &1.06  & 0.71 & 1.64 & 0.86 & 1.05 & 0.45 & 0.36& 0.70& 0.07& 0.45& 0.33 & 0.42 & 0.14 & 0.26 & 0.18 & 0.30 & 0.37 & 0.37 \\ 
    & Fine-tuning & 0.79 &1.19  & 0.59 & 1.82 &0.60  & 1.00 & 0.44 & \textbf{0.32} & \textbf{0.66} &\textbf{ 0.05} & 0.48 & 0.36& 0.36 & \textbf{0.11}  & \textbf{0.25} & 0.19 & \textbf{0.21} & 0.33& \textbf{0.31} \\
    & Reparametrisation & 0.80 & 1.13 & 0.55 &1.67  & 0.55 & 0.94 & 0.53 & 0.42  &0.94 & 0.05& 0.50& 0.44 & 0.38 & 0.19 &0.32 & 0.19 & 0.28 & 0.34 & 0.38 \\
    & Random Backbone & 0.87 & 1.09 & 0.68 & 1.73  &1.04  &1.08 & 0.70 & 0.67 & 1.09 & 0.06 & 0.49 & 0.35 & 0.42 & 0.13 & 0.26 & 0.18 & 0.37 & 0.34 & 0.42\\ 
    & Attention maps & \textbf{0.75} & 1.03 & \textbf{0.55} & 1.65& 0.63 & 0.92 & 0.51  & 0.40 & 0.68  & 0.06  & \textbf{0.40}& 0.30 & 0.38 & 0.15 &0.26  & 0.21 & 0.22 & \textbf{0.31} & 0.32\\
     & ResNet-18 & 0.98 &0.98 & 0.86 & \textbf{1.37} & 0.90  & 1.02 & 0.56 & 0.44 & 0.90& 0.10& 0.47& 0.33 & 0.30 & 0.13 & 0.33 & \textbf{0.15} & 0.49 & 0.60 &  0.40\\
    &  Adaptation layers  &  0.88 & \textbf{0.98} & 0.57 & 1.55 & \textbf{0.55} & \textbf{0.91}& \textbf{0.41} &  0.37 & 0.70 & 0.08 & 0.44 & {\bf 0.30} & \textbf{0.29} & 0.14 & 0.26 & 0.16 & 0.24& 0.39 & 0.32\\
	\hline
\end{tabular}}
\endgroup
\end{center}
\end{table*}

\subsection{Databases}
We evaluate our models on three benchmark databases - FERA 2015 \cite{valstar2015},  DISFA \cite{mavadati2013} and FERA 2017 \cite{valstar2017}. All these datasets contain a set of videos each showing an individual responding to emotion-elicitation tasks. 

\noindent \textbf{FERA2015}~\cite{valstar2015}: The corpus of the FERA2015 challenge is based on the BP4D dataset~\cite{zhang2014}, which is composed of $41$ subjects performing $8$ tasks, plus an extra test set of $20$ subjects, performing additional tasks. The original corpus was released as part of the training and development partitions of the FERA 2015 challenge, whereas the test set, which is not publicly available, was used to rank participants. The training and development partitions are split into $21$ and $20$ subjects. In total, there are $328$ videos corresponding to the training/validation partitions, and $160$ videos corresponding to the test partition. In this paper, we use the official partitions, and report results on both the validation and test set. Given that the test set is not publicly available, we compare our results with those of the challenge winners. All partitions are annotated with $5$ Action Units intensity levels. The training set comprises $\sim 75k$ frames, whereas the validation and test set contain $\sim 69k$ and $\sim 76K$ frames, respectively. \\ 
\noindent \textbf{DISFA}~\cite{mavadati2013}: The DISFA dataset contains video recordings of 27 subjects while watching Youtube videos. Each clip is $\sim4$ length, and has been manually annotated with the intensity of $12$ AU. Given that no official partitions are defined for DISFA, we follow existing works and perform a three-fold cross validation evaluation, where we train a model for each fold, and we report the ICC measured on the aggregated predictions for the whole dataset, each returned by its corresponding model. \\
\noindent \textbf{FERA2017}~\cite{valstar2017}: The FERA2017 corpus extended that of the FERA2015, by augmenting the existing videos with 3D models that are synthesised in $9$ different views. The FERA2017 incorporates in its official training and development partitions the subjects from the test set of the FERA2015 challenge. An additional set of $20$ videos was added as the official test, emanating from the BP4D+ dataset~\cite{zhang2016}. This dataset poses a great challenge in multi-view facial expression recognition, which we prove can be efficiently performed with a computationally simple model. The AU intensity annotations were extended to cover a total number of $7$. For FERA 2017, we use again the official partitions and we report the predictions against the corresponding annotations. FERA 2017 contains roughly $\sim 1.29M$ frames for training, $\sim 696k$ frames for validation, and $\sim 363k$ for testing. \\

\subsection{Evaluation metrics}

We use standard error measures to evaluate AU intensity estimation models. The first measure is the intra-class correlation  (ICC(3,1),  \cite{shrout79}),  commonly used  in  behavioural  sciences  to  measure  agreement  between annotators, and used to rank participants in the FERA challenges.  The  second  measure  is  the  mean  squared  error (MSE) mainly used for prediction problems.

\subsection{Set-up} All experiments are carried out using the Pytorch library for Python \cite{paszke2017}. The adaptation layers along with all versions of the AU estimation network (AU-NET) are trained from scratch with Adam optimiser \cite{kingma2014} and batch size $48$. The weight decay is set to $10^{-6}$ and momentum to $0.9$. We additionally use cosine annealing scheduler with step $5$. Note that during training, FAN weights remain frozen. The ground-truth target heatmaps are generated according to the method described in Sec.~\ref{sec:original}. For BP4D and DISFA, we use the landmarks extracted from the publicly available code of iCCR~\cite{sanchez2016, sanchez2017}, whereas for FERA2017 we used the official implementation of FAN~\cite{bulat2017}. These landmarks are used to define the heatmaps for training. Note that the landmarks are not needed at test time. The facial images are then tightly cropped to $256 \times 256$ resolution to be passed through the corresponding networks. In addition, we use some random augmentation, consiting of flipping, rotation (from $-30$\textdegree to $30$\textdegree),  color jittering, scale noise (from $0.8$ to $1.2$) and random occlusion. In order to ensure a fair comparison, we re-implemented and re-evaluated the Heatmap Regression model proposed in Sec.~\ref{sec:original} and \cite{sanchez2018}. Our new results account for the stronger augmentation and training strategy applied herein.

%% file: content/inhouse_evaluation.tex
\section{In-house evaluation}

\label{sec:evaluation}

\begin{table}[htbp]
\caption{Evaluation of different incremental learning methods on FERA 2017 development partition. Bold numbers indicate best performance.}
\label{table:evaluation_fera2017}
\begin{center}

\resizebox{0.45\textwidth}{!}{
\begin{tabular}{|p{0.001\textwidth}>{\raggedleft} l|*{7}{c}|c|}
  \hline
  \rowcolor{Gray} & Dataset & \multicolumn{8}{c|}{FERA2017}\\
  \rowcolor{Gray}
		& AU & \multicolumn{1}{c}{1} & \multicolumn{1}{c}{4}& \multicolumn{1}{c}{6} & \multicolumn{1}{c}{10} & \multicolumn{1}{c}{12} & \multicolumn{1}{c}{14} & \multicolumn{1}{c|}{17} & Avg. \\
	\hline \hline
	\multirow{6}{*}{\rotatebox{90}{\noindent ICC}} 
    & Scratch&  0.46 & {\bf 0.49} &0.77& {\bf 0.81} & {\bf 0.87}&0.48 & {\bf 0.50}& 0.63 \\ 
    & Fine-tuning  & 0.45 & 0.42& {\bf 0.80} & 0.80 & 0.83 &{\bf 0.63} & 0.48 &0.63\\ 
    & Reparametrisation & 0.45& 0.44& 0.78& 0.80 & 0.86 & 0.56 & 0.46 & 0.62\\
    & Random Backbone  & 0.47 & 0.39 & 0.78 & 0.80& 0.85 & 0.47 & 0.45 & 0.60 \\  
    & Attention maps & 0.52 & 0.44 & 0.77 & 0.79& 0.86 & 0.57 & 0.49 &0.63 \\ 
     & ResNet-18  &  0.46 & 0.45 & 0.76& 0.78 & 0.85 & 0.60 & 0.46 & 0.62 \\  
    & Adaptation layers   & {\bf 0.54} & 0.42 &  0.79 &  0.80 & 0.86 & 0.59&  0.49 & \bf{0.64} \\ 
    \hline \hline
	\multirow{6}{0.1cm}{\rotatebox{90}{MSE}} 
    & Scratch &  0.46& 0.49 & 0.88 & {\bf 0.80} & {\bf 0.74 }&1.26 & 0.73 & {\bf 0.76}\\ 
    & Fine-tuning & {\bf0.39} & 0.55 & 0.86& 0.83 & 0.79 & {\bf 1.06} & 0.92 & 0.77\\
    & Reparametrisation &0.50 & {\bf 0.46} & 0.86 & 0.84 & 0.81& 1.15 & 0.80 &0.77 \\
    & Random Backbone  & 0.54 & 0.47 & {\bf 0.84} & 0.84 & 0.76 & 1.29& 0.75 &0.78\\ 
    & Attention maps  &  0.44 & 0.56 & 0.90 & 0.88 & 0.77 & 1.15 & 0.69 &0.77\\
     & ResNet-18 & 0.58 & 0.49 & 0.93 & 0.88 & 0.81 & 1.13 & 0.83 &  0.81\\
    &  Adaptation layers  & 0.54 & 0.48 & 0.89 & 0.86 & 0.84 & 1.14 & {\bf 0.67} & 0.78\\
	\hline
\end{tabular}}
\end{center}
\end{table}

\begin{table*}[h!]
\caption{Intensity estimation results on FERA 2015 development set and DISFA. (*) Indicates results reported the references. Bold numbers indicate best performance. $\dagger$ indicates in-house reproduced results \vspace{-15pt}}
\label{au_results2}

\begin{center}
\begingroup
\setlength{\tabcolsep}{5.5pt}
\resizebox{\textwidth}{!}{%
\begin{tabular}{|p{0.001\textwidth}>{\raggedleft} l|*{5}{c}|c|*{12}{c}|c|}
  \hline
  \rowcolor{Gray} & Dataset & \multicolumn{6}{c|}{FERA2015} & \multicolumn{13}{c|}{DISFA}\\
  \rowcolor{Gray}
		& AU	& \multicolumn{1}{c}{6} & \multicolumn{1}{c}{10} & \multicolumn{1}{c}{12} & \multicolumn{1}{c}{14} & \multicolumn{1}{c|}{17} & Avg. & \multicolumn{1}{c}{1} & \multicolumn{1}{c}{2} & \multicolumn{1}{c}{4} & \multicolumn{1}{c}{5} & \multicolumn{1}{c}{6} & \multicolumn{1}{c}{9} & \multicolumn{1}{c}{12} & \multicolumn{1}{c}{15} & \multicolumn{1}{c}{17} & \multicolumn{1}{c}{20} & \multicolumn{1}{c}{25} & \multicolumn{1}{c|}{26} & Avg.\\
	\hline \hline
	\multirow{7}{*}{\rotatebox{90}{\noindent ICC}}
    & 2DC \cite{tran2017}* & 0.76 & 0.71 & 0.85 & 0.45 & 0.53 & 0.66 & \textbf{0.70} & \textbf{0.55} & 0.69& 0.05 &\textbf{0.59} & \textbf{0.57} & \textbf{0.88} & 0.32 & 0.10 & 0.08 & 0.90 & 0.50 & 0.50 \\ 
    & CCNN-IT \cite{walecki17}* & 0.75 & 0.69 & 0.86 & 0.40 & 0.45 & 0.63 & - & -& -& -& -& - & - & - & - & - & - & - & - \\
    & VGP-AE \cite{eleftheriadis2016}* & 0.75 & 0.66 & 0.88 & 0.47 & 0.49 & 0.65 & 0.48 & 0.47& 0.62& 0.19& 0.50& 0.42 & 0.80 & 0.19 & 0.36 & 0.15 & 0.84 & 0.53 & 0.46\\
     & HR \cite{sanchez2018}* &  0.79 & {\bf 0.80} & \textbf{0.86} & {\bf 0.54} & 0.43 &  0.68 & - & -& -& -& -& - & - & - & - & - & - & - & -\\
     & HR $\dagger$ &  0.75 & 0.77 &0.85  &0.48  & 0.59 & 0.69 & 0.59 & {\bf 0.56} & 0.74 & 0.36 & 0.53& 0.52 & 0.78 & 0.40 & {\bf 0.38} & 0.13 & 0.91 & 0.53  & 0.53 \\ 
     &  ResNet-18 & 0.74 & 0.75 & 0.85 & 0.40 & 0.52 & 0.65 & 0.41 & 0.35& 0.63& 0.32& 0.49& 0.48 & 0.83 & 0.31 & 0.32 & 0.20 & 0.84 & 0.50& 0.47 \\ 
    & Ours & \textbf{0.79} & 0.76 & 0.84 & 0.52 & \textbf{0.67} & \textbf{0.72} & 0.56 & 0.52 & \textbf{0.75} & \textbf{0.42} & 0.51 & 0.55 & 0.82 & \textbf{0.55} & 0.37 & \textbf{0.21} & \textbf{0.93} & \textbf{0.62} & \textbf{0.57}  \\ 
    \hline \hline
	\multirow{7}{0.1cm}{\rotatebox{90}{MSE}} 
    & 2DC \cite{tran2017}* & \textbf{0.75} & 1.02 & 0.66 & 1.44 & 0.88 & 0.95 & \textbf{0.32} & 0.39& \textbf{0.53}& 0.26& \textbf{0.43} & 0.30 & \textbf{0.25} & 0.27 & 0.61 & 0.18 & 0.37 & 0.55 & 0.37 \\
    & CCNN-IT \cite{walecki17}* & 1.23 & 1.69 & 0.98 & 2.72 & 1.17 & 1.57  & - & -& -& -& -& - & - & - & - & - & - & - & - \\
    & VGP-AE \cite{eleftheriadis2016}* & 0.82 & 1.28 & 0.70 &  1.43 &  0.77 & 1.00 & 0.51 & \textbf{0.32}& 1.13& 0.08& 0.56& 0.31 & 0.47 & 0.20 & 0.28 &0.16 & 0.49 & 0.44 & 0.41 \\
    & HR \cite{sanchez2018} & 0.77 & \textbf{0.92} & 0.65 & 1.57 &  0.77 &  0.94 & - & -& -& -& -& - & - & - & - & - & - & - & - \\
    & HR $\dagger$  & 1.00 &1.06  & 0.71 & 1.64 & 0.86 & 1.05 & 0.45 & 0.36& 0.70& \textbf{0.07}& 0.45& 0.33 & 0.42 & 0.14 & 0.26 & 0.18 & 0.30 & {\bf 0.37} & 0.37 \\ 
    &ResNet-18 & 0.98 & 0.98 & 0.86 & \textbf{1.37} & 0.90  & 1.02 & 0.56 & 0.44 & 0.90& 0.10& 0.47& 0.33 & 0.30 & \textbf{0.13}& 0.33 & \textbf{0.15} & 0.49 & 0.60 &  0.40 \\ 
    &  Ours  &  0.88 & 0.98 & \textbf{0.57} & 1.55 & \textbf{0.55} & \textbf{0.91}& 0.41 &  0.37 & 0.70 & 0.08 & 0.44 & {\bf 0.30} & 0.29 & 0.14 & \textbf{0.26} & 0.16 & \textbf{0.24} & 0.39 & \textbf{0.32}\\
	\hline
\end{tabular}
}
\endgroup
\end{center}
\end{table*}

\noindent In this Section we analyse and evaluate all our proposed approaches for AU prediction. To do so, we experimentally evaluate the performance of all our methods using ICC score and Mean Square Error on all three aforementioned datasets (FERA2015, FERA2017 and DISFA). Results on FERA 2015 and DISFA are shown in Table~\ref{table:evaluation_fera2015_disfa}, while results on FERA 2017 are on Table~\ref{table:evaluation_fera2017}. Also, to further allow for a comprehensive evaluation of the strengths and weaknesses of each of the methods we include a thorough evaluation of their complexity, including the capacity of each model, the number of floating point operations per second, and the average time per forward pass. The summary of complexity is illustrated in Table~\ref{table: technical_comparisons}, along with the performance each method reports on the validation set of BP4D, often used as the referent benchmark for comparison in existing works. In addition to our proposed methods, we include a strong baseline based on a ResNet-18 \cite{he2016}, which is a rather deep network of approximately $11M$ parameters. For the task of AU intensity estimation, we modify the last layer to generate predictions that match the number of Action Units. Then, we simply regress AU intensity levels. All models are trained under the same training configuration, i.e same learning rate, optimisers and augmentation.

\subsection{Heatmap regression vs. regression}
\label{ss:heatmap_regression2regression}
We approach the  task AU intensity estimation from a geometric perspective. We train our models in a Heatmap Regression fashion which allows us to jointly localise the AUs and estimate their intensity. This approach is simple and straightforward: an image is passed forward to the AU-net that generates a set of heatmaps from which simple retrieval of the maximum value gives AUs predictions. To evaluate the impact of heatmap regression, we train a model to simply regress AU intensity levels rather than regressing heatmaps. This model is ResNet-18 which also serves as our baseline. As shown in Table~\ref{table:evaluation_fera2015_disfa} and Table~\ref{table:evaluation_fera2017}, Heatmap Regression significantly outperforms ResNet-18 both in terms of ICC score and Mean Square Error in all three datasets. Notably, our method achieves better results than direct regression with a model of much less capacity.

\begin{table}[htb!]
\caption{Computation complexity of the proposed methods. \vspace{-15pt}}
\label{table: technical_comparisons}
\begin{center}
\begingroup
\setlength{\tabcolsep}{1.2pt}
\resizebox{0.50\textwidth}{!}{%
\begin{tabular}{|p{0.001\textwidth}>{\raggedleft} l|*{3}{c}|*{3}{c}|c|}
  \hline
  \rowcolor{Gray} & Method & \multicolumn{3}{c|}{Complexity} & \multicolumn{3}{c|}{ ICC }\\
  \rowcolor{Gray} &  & \multicolumn{1}{c}{Num. of parameters} & \multicolumn{1}{c}{flops} & \multicolumn{1}{c|}{secs/img } & \multicolumn{1}{c}{ FERA 2015}& \multicolumn{1}{c}{ DISFA} & \multicolumn{1}{c|}{ FERA 2017} \\
	\hline \hline
	 	\multirow{7}{*}{}& ResNet-18 & 11.17M &4.75G &0.004 & 0.65 &  0.47&0.62 \\
	 	& Scratch  & 1.65M  & 5.21G & 0.012 & 0.69 & 0.53 &0.63\\
	 	 & Fine-tuning  & 1.65M & 5.21G  & 0.012 &0.71 & 0.56& 0.63\\
	 	 & Reparametrisation  &  1.79M &  5.21G  &0.016  & 0.70 & 0.54& 0.62 \\
    	& Attention maps  & 2.65M& 6.32G & 0.021 & 0.71 & 0.57 & 0.63\\
    & Adaptation layers &  2.65M & 6.32G & 0.018  &0.72 &0.57 &0.64 \\
    \hline 
\end{tabular}
}
\endgroup
\end{center}
\end{table}

\subsection{Heatmap regression  vs transfer learning}
\label{ss:heatmapvsincremental}
We extend our approach of AU intensity estimation through Heatmap Regression by proposing methods that absorb knowledge of a pre-existing network for face alignment. Thus, we further investigate whether our four different methods, i.e fine-tuning, adaptation layers, reparametrisation and attention maps, that leverage information from FAN, benefit the task of Heatmap Regression for AU estimation. To evaluate their impact, we first train all methods under the very same training scenario, i.e same data augmentation, same training configuration etc., and test the performance of each in terms of ICC and MSE score. Then, we further evaluate their computational requirements with regards results in Table~\ref{table: technical_comparisons}. We refer to the Heatmap Regression method presented in Sec.~\ref{sec:original} as trained from \textit{scratch}, and to each of the transfer learning methods by their corresponding technique.

As shown in Table \ref{table:evaluation_fera2015_disfa} and Table \ref{table:evaluation_fera2017} transfer learning improves over training from scratch in almost all three datasets in terms of both ICC and MSE. For FERA 2015, all transfer learning methods yield an ICC score that ranges between $0.70$ and $0.72$, while the model trained from scratch achieves an ICC score of $0.69$. The same behaviour is observed for DISFA, where transfer learning improves over training from scratch, with ICC scores ranging from $0.54$ to $0.57$ for the former, vs. the $0.53$ given by the latter. However, we observe that for FERA 2017 both training from scratch and applying transfer learning appear to deliver similar results, which we attribute to the fact that indeed FERA 2017 is a large-scale dataset that includes a large variety of poses. Under such a large pool of videos, training the model from scratch suffices to yield competitive results. 

Regarding complexity (see Table~\ref{table: technical_comparisons}), we observe that all our methods deploy models with a small number of parameters that roughly range between $1.6M$ to $2.65M$. The transition from having a model trained from scratch to transfer learning requires no extra parameters when fine-tuning, and a negligible number of extra parameters for the reparametrisation approach. The use of adaptation layers and attention maps incur in only an extra $1M$ number of parameters. We can observe that this increase is negligible compared to the original Hourglass of \cite{bulat2017}, which comprises $3M$. Similarly, the number of slightly increase in the case of adaptation layers and attention maps.  

\begin{table*}[htp!]
\caption{Intensity estimation results on FERA 2017 development and test set. (*) Indicates results reported the reference. Bold numbers indicate best performance. $\dagger$ indicates in-house reproduced results \vspace{-15pt}}
\label{fera2017}
\begin{center}
\begingroup
\setlength{\tabcolsep}{5.5pt}
\resizebox{\textwidth}{!}{%
\begin{tabular}{|p{0.001\textwidth}>{\raggedleft} l|*{7}{c}|c|*{7}{c}|c|}
  \hline
  \rowcolor{Gray} & Dataset & \multicolumn{8}{c|}{FERA 2017 (development set)} & \multicolumn{8}{c|}{FERA 2017 (test set) }\\
  \rowcolor{Gray}
		& AU & \multicolumn{1}{c}{1} & \multicolumn{1}{c}{4 }& \multicolumn{1}{c}{6} & \multicolumn{1}{c}{10} & \multicolumn{1}{c}{12} & \multicolumn{1}{c}{14} & \multicolumn{1}{c|}{17} & Avg. & \multicolumn{1}{c}{1} & \multicolumn{1}{c}{4 }& \multicolumn{1}{c}{6} & \multicolumn{1}{c}{10} & \multicolumn{1}{c}{12} & \multicolumn{1}{c}{14} & \multicolumn{1}{c|}{17} & Avg.\\
	\hline \hline
	\multirow{7}{*}{\rotatebox{90}{\noindent ICC}}
& AUMPNet\cite{batista2017}* & 0.38 & 0.23 & 0.66 & 0.69 & 0.76 & 0.46& 0.33& 0.50& 0.23 & 0.06 & 0.70 & 0.71 & 0.73 & 0.10& 0.26& 0.40 \\
& SVR \cite{amirian2017}* & 0.18 & 0.13 & 0.50 & 0.65 & 0.63 & 0.48 & 0.20 & 0.40  & 0.17 & 0.02 & 0.51 & 0.59 & 0.62 & -0.03& 0.19& 0.30\\
& MTask (Inv)\cite{zhou2017}* & 0.44 & 0.25 & 0.72 & 0.77 & 0.80 & 0.55 & 0.34 & 0.55  & - & - & - & - & - & -& -& - \\ 
& MTask (Dep)\cite{zhou2017}* & 0.54 & 0.41 & 0.70 & 0.78 & 0.82 & 0.50 & 0.45 & 0.60  & 0.31 & 0.15 & 0.67 & 0.74 & 0.79 & 0.15 & 0.32 & 0.45\\ 
& HR \cite{sanchez2018}$\dagger$ &  0.46 & {\bf 0.49} &0.77& 0.81 & {\bf 0.87}&0.48 & {\bf 0.50 }& 0.63 &  0.38 &  0.20 & 0.77 &{\bf 0.79}& 0.82
& {\bf 0.19} & 0.40& {\bf0.51} \\    
& ResNet-18  &  0.46 & 0.45 & 0.76& 0.78 & 0.85 &{\bf 0.60}& 0.46 & 0.62  &0.23 & 0.08 & 0.75 &0.77 & 0.79  & 0.17 & 0.36 & 0.45 \\
& Ours & {\bf 0.54} & 0.42 & {\bf 0.79} & {\bf 0.80} & 0.86 & 0.59&  0.49 & \bf{0.64}  & {\bf 0.31}  & {\bf 0.23} & {\bf 0.78} & 0.78 & {\bf 0.82}&0.18 & {\bf 0.40}& 0.50 \\ 
\hline \hline
\multirow{7}{0.1cm}{\rotatebox{90}{RMSE}} 
    & AUMPNet\cite{batista2017}* & - & - & - & - & - & -& -& -  & - & - & - & - & - & -& -& -  \\
    &SVR\cite{amirian2017}*& {\bf0.42} & {\bf0.47} &  1.79 & 1.46& 1.59 & 2.07 & 0.95 & 1.25  &0.55& {\bf 0.32} & 1.29  & 1.19 & 1.21& 1.53 & 0.70 &0.97 \\
    & MTask (Inv)\cite{zhou2017}* &0.55 & 0.56& 0.99& 0.88 & 0.91 & 1.14& 0.87& 0.84 & - & - & - & - & - & -& -& - \\ 
    & MTask (Dep)\cite{zhou2017}* &0.52& 0.57 & 0.99 & 0.95 & 0.83 & 1.17& 0.75 &0.82 & 0.74 & 0.47 & 0.97 & 1.05 & {\bf 0.87} &{\bf  1.21} & 0.85 & 0.88 \\
	& HR \cite{sanchez2018}$\dagger$&  0.46& 0.49 & {\bf 0.88} & {\bf 0.80} & {\bf 0.74 }&1.26 &{\bf 0.73} & {\bf 0.76} &  {\bf 0.55} & 0.49 & {\bf 0.96}  & {\bf 0.88} & 0.97 &
       1.27& 0.76 &{\bf 0.84}  \\   
    	& ResNet-18 & 0.58 & 0.49 & 0.93 & 0.88 & 0.81 &{\bf  1.13} & 0.83 &  0.81 &	0.95 & 0.53 & 1.02 & 0.92 & 1.03& 1.39& 0.90 & 0.96\\
	& Ours	& 0.54 & 0.48 & 0.89 & 0.86 & 0.84 & 1.14 & 0.67 & 0.78 &  0.89 & 0.42 & 1.00 & 0.98& 1.06 & 1.63 & {\bf 0.69} & 0.95 \\
	\hline
\end{tabular}
}
\endgroup
\end{center}
\end{table*}

\subsection{Comparison between transfer learning methods}
\label{ss:incremntal_comparison}
We now turn our analysis to the comparison between each of the proposed methods for transfer learning. While the discussed approaches deliver state of the art results (see Sec.~\ref{ss:experiments}) it is worth discussing the pros and cons of each, according to their performance and complexity. 

The first proposed approach to transfer learning, that of \textit{fine-tuning}, is undoubtedly the simplest in terms of complexity, which matches that of training the network from scratch. We observe that fine-tuning brings a considerable gain in performance w.r.t. training from scratch, especially in DISFA, which shows to be an effective and efficient way to transfer learning. 

In the same line, we can observe that, while the \textit{reparametrisation} approach results in an even more efficient approach to that of fine-tuning (much less number of learnable parameters, with same inference complexity), its performance is slightly worse. 

Arguably, the best gain in performance comes from the adaptation layers and the attention maps. While the latter includes an extra step to convert the facial landmark heatmaps into AU-attention maps, the complexity can be said to be the same. However, we observe that using the adaptation layer directly from the heatmaps returned by the FAN outperforms the results given by the attention maps. We attribute this to the fact that the detected heatmaps provide some confidence, which can be more effectively used by the network to automatically infer the attention.

In summary, while fine-tuning and reparametrisation seem to be the methods with the least complexity, the \textit{adaptation layers} method yields the best performance. However, it is worth noting that, regardless this method being the most complex from the proposed ones, its complexity and number of parameters compared to those of the Resnet-18 suggest this as an efficient method for AU localisation, and thus we choose it for comparison w.r.t. state of the art works.

\subsection{Core task with random weights}
\label{ss:random}
The proposed transfer learning methods depart from a core network pre-trained for the task of Face Alignment, and include an extra set of learnable weights to perform the target task of AU localisation and intensity estimation. With the great success of the Heatmap Regression method proposed in Sec.~\ref{sec:original}, it is natural to explore whether the gain in performance comes from having a more constrained network, or from the actual features inherited from the FAN network. To validate that the contribution of the transfer learning methods does not come from the little capacity added to the core network, we study the performance of using the adaptation layers using as a core network a FAN-like network that is initialised with random weights, and that remains frozen with these randomly initialised weights. The results of this study are those referred to as \textit{random backbone} in Table~\ref{table:evaluation_fera2015_disfa} and Table~\ref{table:evaluation_fera2017}. It can be seen that, while learning only the extra network still produces competitive results, having the rich representations given by the FAN is crucial to achieve state of the art results. Note that, despite the network receiving the features from a randomly initialised FAN, the generated features after the $\mathtt{conv4}$ layer are still conditioned to the input image, through some fixed random non-linear projections.

\subsection{Fine-tuning after transfer learning}
In addition to the aforementioned studies, we also explored an alternative approach that consists of fine-tuning the whole pipeline after the transfer learning step. In particular, we unfreeze the FAN network and we fine-tune the whole pipeline. We, however, observed no improvement in the performance. 

%% file: content/comparison_soa.tex
\begin{figure*}[htp!]
    \centering
    \includegraphics[width=1\textwidth]{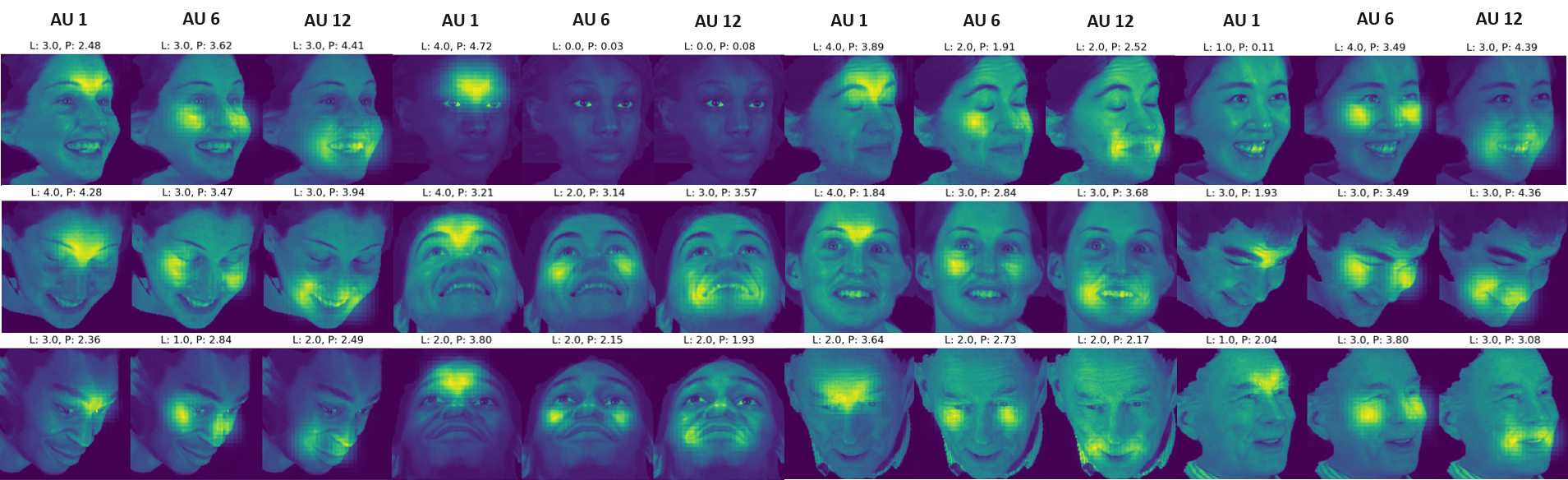}
    \caption{Visual examples of AU localisation for three different AUs on FERA 2017. We superimpose to the input images the regressed heatmaps (upsampled to match the input resolution). The magnitude and the peak of each Gaussian is corresponding to the predicted AU intensity which is also shown at the top of each image along with the corresponding label. }
    \label{fig:localisation}
\end{figure*}
\section{Comparison with state of the art}
\label{ss:experiments}

In this Section, we report the results of our proposed approach w.r.t. state of the art results in both the validation and test partitions of FERA 2015 and FERA 2017, as well as after the 3-fold cross-validation experiment on DISFA. Similarly to Sec.~\ref{sec:evaluation}, we use as a baseline a ResNet-18 \cite{he2016} which is trained to directly regress AU intensity levels. In addition to that, we report the results of the network trained from scratch, herein referred simply as HR. Finally, for the sake of clarity, we report the results of the best performing method from the proposed transfer learning approaches, that of the \textit{Adaptation Layers} (Sec.~\ref{ss:feature_injection}), herein simply referred to as \textit{Ours}.

\subsection{FERA 2015 dataset}
\label{ss:fera2015}
The results for FERA 2015 - Development are shown in Table~\ref{au_results2}, whereas those regarding the test partition are shown in Table~\ref{resultsfera15test}. It is important to recall that, given that FERA 2015 is not publicly available, current works report on the development set, hence the lack of up-to-date results on the test partition. Despite the recent advances and the improved results on the development set, our method outperforms state of the art results by a considerable margin. We observe that the transfer learning approach results crucial to attain a new state of the art result in both partitions, proving the effectiveness of such approach when working with small datasets. 

\begin{table}[htb!]
\caption{The intensity estimation results on FERA 2015 test partition. Comparison with the challenge winners. (*) Indicates results reported by the references. Bold numbers indicate best performance. \vspace{-15pt}}
\label{resultsfera15test}
\begin{center}
\begingroup
\setlength{\tabcolsep}{5.2pt}
\resizebox{0.45\textwidth}{!}{%
\begin{tabular}{|p{0.001\textwidth}>{\raggedleft} l|*{5}{c}|c|}
  \hline
  \rowcolor{Gray} & Dataset & \multicolumn{6}{c|}{FERA2015}\\
  \rowcolor{Gray}
		& AU & \multicolumn{1}{c}{6} & \multicolumn{1}{c}{10} & \multicolumn{1}{c}{12} & \multicolumn{1}{c}{14} & \multicolumn{1}{c|}{17} & \multicolumn{1}{|c|}{Avg.}\\
	\hline \hline
	 	\multirow{5}{*}{\rotatebox{90}{\noindent ICC}} 
    &ISIR \cite{nicole15}*&  0.79 & 0.80 & \textbf{0.86} & 0.71 & 0.44  & 0.72\\
    &CDL \cite{Baltrusaitis2015}*&  0.69 & 0.73 & 0.83 & 0.50 & 0.37  & 0.62\\ 
        & HR-Scratch &  0.79& 0.81 & 0.82 & 0.60 & 0.41  & 0.69 \\
        & ResNet-18 & 0.74 & 0.75 & 0.85 & 0.40 & 0.52 & 0.65 \\
    & Ours &  \textbf{0.82} & \textbf{0.82} & 0.80 & \textbf{0.71} & \textbf{0.50}  & \textbf{0.73}\\

    \hline \hline
	\multirow{5}{0.1cm}{\rotatebox{90}{MSE}} 
      &ISIR \cite{nicole15}*&  0.83 & 0.80 & \textbf{0.62} & 1.14 & 0.84  & 0.85\\
    &CDL \cite{Baltrusaitis2015}*&  - & - & - & - & -  & -\\ 
      & HR-Scratch & 0.96 & 0.97 & 0.82& 1.08& 1.10 & 0.99\\
      & ResNet-18 & 0.98 & 0.98 & 0.86 & 1.37 & 0.90  & 1.02\\
    & Ours &  \textbf{0.68} & \textbf{0.80} & 0.79 & \textbf{0.98} & \textbf{0.64} & \textbf{0.78} \\
    \hline
\end{tabular}
}
\endgroup
\end{center}
\end{table}

\subsection{DISFA dataset}
\label{ss:disfa}
We report the results of the 3-fold cross-validation experiment on DISFA in Table~\ref{au_results2}. We can observe that both Heatmap Regression and transfer learning outperform existing methods in such a challenging dataset. We attribute this gain to the fact that localising the AUs is more effective than resorting to complex Autoencoder networks such as the one proposed in the 2DC~\cite{tran2017}. With Heatmap Regression, the network returns a structured representation that already captures the AU dependencies in a geometric way, and thus no additional dependencies need to be learned.   

\subsection{FERA2017 dataset}
\label{ss:fera2017}

Results on FERA 2017 validation and test set are given in Table~\ref{fera2017}. Note that for FERA 2017 we choose to report the Root Mean Squared Error as this was the measure of choice for the challenge. Our approach achieves an ICC score of $0.64$ on the validation set which is by $7\%$ better than the ICC score of FERA 2017 challenge winners, that attained an ICC score of $0.60$. Similarly, in terms of RMSE score our method method outperforms challenge winners by a $5\% $ margin. The same pattern is also found on the test set, where our method reports an ICC score of $0.50$, which surpasses $0.45$ ICC reported by the challenge winners. We can also observe that both HR-Scratch and our transfer learning approach deliver similar results, which we attribute to the fact that FERA 2017 is already a large-scale dataset. 

\noindent \textbf{Qualitative evaluation}: In addition to the reported results, we show the capabilities of our method to actually infer both the location and the intensity of Action Units in Fig.~\ref{fig:localisation}. We observe that for FERA 2017, that spans a large set of poses, our method is capable of estimating the location and intensity accurately, thus proving the efficacy of Heatmap Regression for the task of AU intensity estimation.